\documentclass[journal]{IEEEtran}
\ifCLASSINFOpdf
\else
\fi

\usepackage{times}
\usepackage{epsfig}
\usepackage{graphicx}
\usepackage{amsmath}
\usepackage{amssymb}
\usepackage{multirow}

\usepackage{graphicx}  
\usepackage{booktabs}
\usepackage{float} 
\usepackage{booktabs}
\usepackage{setspace}
\usepackage{color}
\usepackage[dvipsnames]{xcolor}
\usepackage{algorithm} 
\usepackage{pdfpages}
\usepackage{subfigure}
\usepackage{tabularx}

\usepackage{array}
\makeatletter
\newcommand{\thickhline}{%
	\noalign {\ifnum 0=`}\fi \hrule height 1pt
	\futurelet \reserved@a \@xhline
}
\newcolumntype{"}{@{\vrule width 1pt}}
\makeatother

\makeatletter  
\newcommand*\bigcdot{\mathpalette\bigcdot@{.5}}
\newcommand*\bigcdot@[2]{\mathbin{\vcenter{\hbox{\scalebox{#2}{$\m@th#1\bullet$}}}}}
\makeatother
\begin{document}
%
\title{Unified and Dynamic Graph for Temporal Character Grouping in Long Videos}
%
%
%

\author{Xiujun Shu$^{\dagger}$,	
        Wei Wen$^{\dagger}$,
        Liangsheng Xu$^{\dagger}$, 
        Ruizhi Qiao$^{*}$,
        Taian Guo,
        Hanjun Li,\\
        Bei Gan,
        Xiao Wang,
        Xing Sun 
\thanks{$^{\dagger}$ Equal Contribution}  
\thanks{Xiujun Shu, Wei Wen, Liangsheng Xu, Ruizhi Qiao, Taian Guo, Hanjun Li, Bei Gan, and sun xin are with Tencent YouTu Lab, Shenzhen, China. (e-mail: \{xiujunshu, jawnrwen, seanlsxu, ruizhiqiao, taianguo, hanjunli, stylegan and winfredsun\}@tencent.com)}
\thanks{Xiao Wang is with the School of Computer Science and Technology, Anhui University, Hefei 230601, China. (e-mail: wangxiaocvpr@foxmail.com)}    
}

%
%

\markboth{Journal of \LaTeX\ Class Files,~Vol.~14, No.~8, August~2023}%
{Shell \MakeLowercase{\textit{et al.}}: Bare Demo of IEEEtran.cls for IEEE Journals}
%



\maketitle

\begin{abstract}
Video temporal character grouping locates appearing moments of major characters within a video according to their identities. To this end, recent works have evolved from unsupervised clustering to graph-based supervised clustering. However, graph methods are built upon the premise of fixed affinity graphs, bringing many inexact connections. Besides, they extract multi-modal features with kinds of models, which are unfriendly to deployment. In this paper, we present a unified and dynamic graph (UniDG) framework for temporal character grouping. This is accomplished firstly by a unified representation network that learns representations of multiple modalities within the same space and still preserves the modality's uniqueness simultaneously. Secondly, we present a dynamic graph clustering where the neighbors of different quantities are dynamically constructed for each node via a cyclic matching strategy, leading to a more reliable affinity graph. Thirdly, a progressive association method is introduced to exploit spatial and temporal contexts among different modalities, allowing multi-modal clustering results to be well fused. As current datasets only provide pre-extracted features, we evaluate our UniDG method on a collected dataset named MTCG, which contains each character's appearing clips of face and body and speaking voice tracks. We also evaluate our key components on existing clustering and retrieval datasets to verify the generalization ability. Experimental results manifest that our method can achieve promising results and outperform several state-of-the-art approaches.

\end{abstract}


%
\IEEEpeerreviewmaketitle

\section{Introduction} \label{introduction}
\IEEEPARstart{W}{ith} the proliferation of video platforms, identity-oriented character grouping~\cite{fitzgibbon2002affine,choi2010automatic,jin2017end} has received substantial popularity from both academia and industries due mostly to its precious value to many crucial applications, \emph{e.g.,} video editing, specific browsing, celebrity collection, and story understanding. Related works in this field have evolved from unsupervised to graph-based supervised clustering, and from single-modal to multi-modal clustering. Early works~\cite{li2016subspace,kalogeiton2020constrained,pei2020efficient} are unsupervised and confined to simply learning from small datasets, \emph{e.g.,} Mice Protein~\cite{lecun1998gradient} and MNIST~\cite{higuera2015self}. Later works ~\cite{2019Learning,chen2021low,yang2021gcn} demonstrated that graph-based supervised clustering can achieve better performance in large-scale datasets, \emph{e.g.,} CASIA~\cite{2014Learning} and MegaFace~\cite{kemelmacher2016megaface}. For example, L-GCN~\cite{wang2019linkage} and GCN-VE~\cite{yang2020learning} leverage graph convolutional network (GCN) to capture the graph context for face clustering. Unfortunately, single modality recalls few positive shots on account of massive faces invisible from the camera. This has recently motivated the community to put more effort into exploring multi-modal cues to achieve higher precision and recall. Brown~\emph{et al}.~\cite{Brown2021FaceBV} have released a multi-modal person-level dataset VPCD for better-locating characters in videos. Nevertheless, only pre-extracted features are provided, making it impossible for the followers to explore more powerful representations for clustering. It plays a decisive role in actively fulfilling satisfactory performance of temporal character grouping, as we demonstrate in this work.

\begin{figure}[t]
	\centering
	\includegraphics[width=\linewidth]{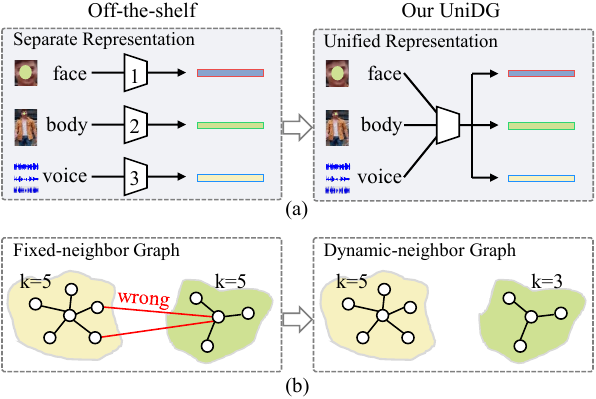} 
\caption{Off-the-shelf video character-grouping methods \emph{v.s.} our UniDG in this paper. (a) We learn representations across modalities in the same space. (b) We construct dynamic neighbors for more reliable affinity graphs.} 
	\label{fig:motivation}
\end{figure}

\begin{figure*}[t]
	\centering
	\includegraphics[width=\linewidth]{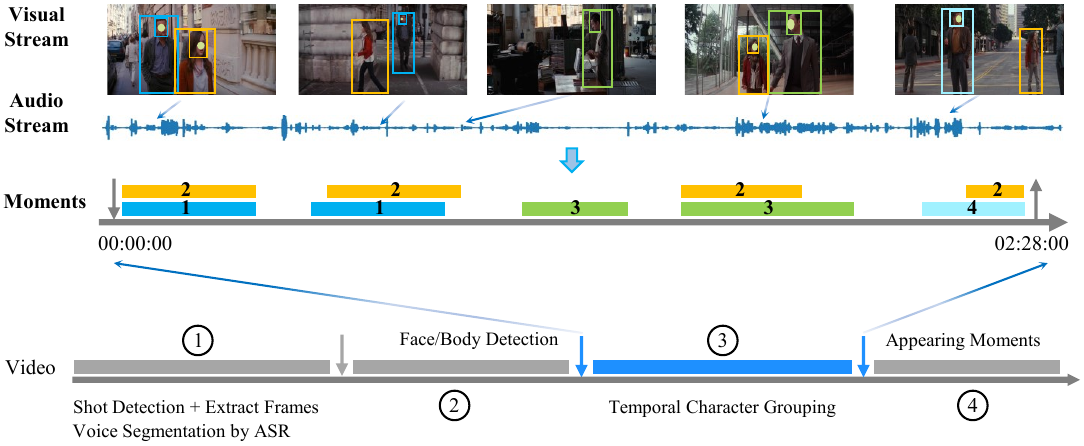} 
        \caption{The definition of multi-modal temporal character grouping in long videos. The whole pipeline in real applications contains four stages and we focus on the stage of temporal character grouping. The long videos usually last tens of minutes to hours. By fully exploring multiple modalities, \emph{i.e.,} face, body, and voice, we can obtain the start and end time of appearing moments for major characters (1, 2, 3, \emph{etc.}) in each video. Multiple modalities can provide complementary cues, \emph{e.g.,} the girl's face in the 2nd image is not visible, but her body can recall corresponding moments. The voices are used to assist us in obtaining more compact grouping results.} 
	\label{fig:definition}
\end{figure*}

In this paper, we devise a unified and dynamic graph (UniDG) framework for temporal character grouping in long videos. As shown in Fig.\,\ref{fig:motivation}, the core idea involves two aspects: \emph{unified representation} and \emph{dynamic graph clustering}. This is inspired by two observations that previous works have ignored: 
1) Video contains multi-modal signals, \emph{e.g.,} face, head, body, voice, gaits, title, and captioning. Previous clustering methods require to deploy of several models to extract features for each modality. This is unfriendly in resource-constrained scenarios. One of our goals is to learn representations of all modalities with a single model that can achieve competitive performance to expert models. 
2) State-of-the-art supervised methods implement graph convolutional network (GCN) on the premise of fixed $k$ neighbors for each node in the constructed affinity graph. This style may lead to many inexact connections, especially in TV shows where characters wear similar clothes or change their clothes to attend different occasions. Another goal of ours is to explore the feasibility of dynamic graph clustering.


To achieve the first goal, we need to answer a key question: \emph{Can unified representation be competitive compared with expert models}? Some works~\cite{li2021align,li2023uni} have conducted some explorations, but how to balance the training of different modalities is still challenging~\cite{ren2022balanced}. A recent work ImageBind~\cite{girdhar2023imagebind} has unified six modalities to the same space, but it focuses on cross-modal semantic alignment and sacrifices the modality uniqueness to some extent. As shown in Fig.\,\ref{fig:motivation}(a), we leverage a single model to learn representations for all modalities and still keep the modality uniqueness. This is achieved by unifying all modalities into the same label space. Through extensive experiments, we demonstrate that unified representation is competitive to separate expert representations.
To achieve the second goal, we construct the affinity graph where different nodes possess different neighbors in quantities, as illustrated in Fig.\,\ref{fig:motivation}(b), achieving more reliable edge connections for better optimization. 
In addition to supervised clustering, performance improvement is also observed when making an extension to unsupervised settings. 
%
After performing clustering upon all modalities, we further propose a progressive association method to conduct temporal character grouping. This method considers the spatial-temporal context of different modalities and achieves significant performance improvement to single-modality. 
To evaluate our method,  we collect a multi-modal temporal character grouping (MTCG) dataset. It provides shot-level raw data and accurate annotations for modalities including face, body, and voice. It is by far the largest multi-modal dataset for temporal character grouping.

The main contributions of this paper can be summarized as the following:
\begin{itemize}
\item We provide a deployment-friendly framework for temporal character grouping. This framework contains representation, clustering, and fusion for modalities in long videos. 


\item  We are the first to explore dynamic graphs for the temporal character grouping, which can be applied to both supervised and unsupervised settings. Besides, we propose a training strategy enabling unified representations to be competitive with separate export representations.




\item We conduct comprehensive experiments on the collected MTCG dataset, typical face clustering datasets, and retrieval datasets. Experimental results validate the effectiveness of our framework.

\end{itemize}

 \begin{figure*}[!t]
	\centering    
    \includegraphics[width=\linewidth]{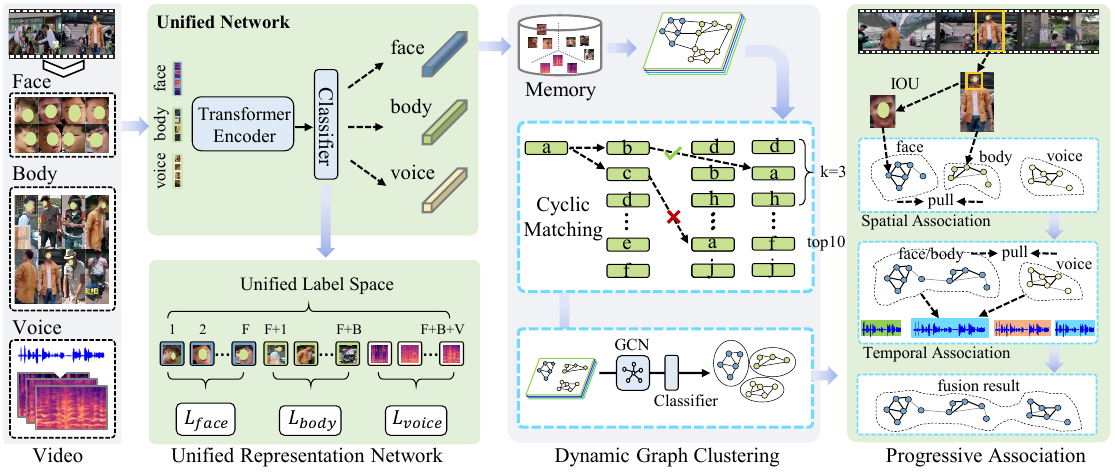}
	\caption{Architecture of the proposed UniDG Framework. 
		The framework consists of three modules: Unified Representation Network, Dynamic Graph Clustering, and Progressive Association. The input consists of three modalities, \emph{i.e.,} face, body, and voice. 
		}	
	\label{fig:framework}
\end{figure*}

\section{Related Work} 
  
\subsection{Representation Learning}
Representation learning targets extract discriminative features, allowing to support of multiple visual tasks. Current methods can be classified into supervised and self-supervised learning. Supervised representation learning, \emph{e.g.,} metric learning~\cite{cakir2019deep},  benefits from large-scale labeled datasets, \emph{e.g.,} ImageNet~\cite{deng2009imagenet}. Its core idea is to pull together samples within the same category in embedding space and push them apart otherwise~\cite{wang2014learning}. However, large-scale labeling is expensive and labor-intensive. In recent years, self-supervised learning has achieved encouraging results~\cite{he2020momentum,zhu2022self,fu2021self}. With various pretext tasks, it can learn underlying representations without using any labels. Benefiting from the powerful modeling of Transformer~\cite{dosovitskiy2020image}, multi-modal and cross-modal pre-training~\cite{zhu2019multi,chen2020uniter,Zhang2021VinVLRV,nie2020deep,feng2022temporal} has been widely studied recently. A recent work ImageBind~\cite{girdhar2023imagebind} has unified six modalities to the same space, which would significantly reduce the memory requirement in real deployment. However, these works focus on the network architecture design or cross-model semantic alignment. They would sacrifice the modality uniqueness to some extent. In this work, we aim to learn unified representation but retain the modality uniqueness simultaneously.

\subsection{Clustering Algorithms} 
Traditional clustering methods are unsupervised, \emph{e.g.,} $k$-Means~\cite{lloyd1982least}, DBSCAN~\cite{ester1996density}, and Spectral Clustering~\cite{fowlkes2004spectral}. Such methods usually depend
on some data assumptions, \emph{e.g.}, assuming that the same class has a cluster center. Consequently, they lack the capability of coping with complicated cluster structures. Later works focus on video face clustering by exploring the spatio-temporal continuity with link constraints~\cite{bauml2013semi,cinbis2011unsupervised,kalogeiton2020constrained,sharma2020clustering,wu2013constrained}. For instance, C1C~\cite{Kalogeiton2020ConstrainedVF} links instances through first NN relations.
Most works solve face clustering with representation or metric learning~\cite{sharma2019video,sharma2020clustering,tapaswi2019video,wu2013simultaneous,ghasedi2017deep}. Some approaches also employ the self-supervised method, {e.g.,} SSIAM~\cite{sharma2019self}, FGG~\cite{rothlingshofer2019self} and CCL~\cite{sharma2020clustering}. Recent works\cite{yang2019learning,yang2020learning,wang2019linkage,wang2022ada} focus on GCN-based supervised learning and have achieved impressive results in large-scale datasets. 
STAR-FC~\cite{shen2021structure} designs a structure-preserved sub-graph sampling strategy. 
HSAN~\cite{liu2022hard} mines the hard samples in contrastive deep graph clustering. However, these methods only utilize faces and ignore other informative cues in videos. A recent work~\cite{Brown2021FaceBV} develops face clustering to multi-modal person clustering. It proposes a multi-stage fusion strategy for multi-modal clustering. This is the most similar work to our research, but it focuses on traditional unsupervised clustering. This work considers both representation and dynamic graph clustering, which play decisive roles in temporal character grouping. 



\section{Methodology} 
 
\subsection{Preliminaries}  

Fig.\,\ref{fig:definition} gives the whole pipeline of using our temporal character grouping in real applications. The pipeline contains four stages and this work focuses on the third stage, \emph{i.e.,} temporal character grouping. Video character grouping locates appearing moments of major characters within a video according to their identities. This task can obtain the start and end time of a person's appearance in the video. As shown in Fig.\,\ref{fig:definition}, this work extends previous single-modal face clustering to multi-modal temporal character grouping. The major difference lies in two aspects: 
1) More modalities. It includes face, body, and voice. Although face and body are both visual modalities, their semantic cues are quite different, thus we regard them as different modalities. 
2) Temporal moments. Previous face clustering only focuses on the clustering algorithm, we aim to locate the appearing moments of major characters. 


In videos, voices are sometimes not synchronized to faces and bodies. For example, a certain face appears but the voice is from another person who does not appear. Besides, there may be a moment when faces and bodies can be seen, but no one is talking. These challenges force us to give a more clear statement that \emph{each cluster includes the clips of a person when he or she appears and does not include moments containing only his or her voices}. The voices are only used to assist us in obtaining more compact clustering results, but do not provide more clips. For example, if the voice of a person appears at a certain moment but he or she is not visible, the moment should not be included in the person's cluster.

 

Video naturally contains information of various modalities. This work considers three modalities, \emph{i.e,} face, body, and voice. All samples of modalities are pre-extracted from the raw videos. The face and body are two sets of images, and the voice is a collection of one-dimensional sound sequences.
%
Assume they are denoted as ${X^f=\left[x^f_{1}, x^f_{2},...,x^f_{M_f}\right]}$ with its label set $L^{f}=\{1,2,...,F\}$, ${X^b=\left[x^b_{1}, x^b_{2} ,...,x^b_{M_b}\right]}$ with its label set $l_{b}=\{1,2,...,B\}$, and ${X^v=\left[x^v_{1}, x^v_{2},...,x^v_{M_v}\right]}$ with its label set $l^{v}=\{1,2,...,V\}$, respectively. $M_f$, $M_b$, $M_v$ denote the sample numbers of three modalities, and $F$, $B$, $V$ denote the category numbers.
Each sample records its spatial and temporal locations in the original video. The objective in this paper is to obtain the appearing moments of each identity by clustering $X^f$, $X^b$ and $X^v$. 
%
For ease of representation, the superscripts $f$, $b$ and $v$ might be dropped from time to time in the following.

%
%

%
The framework of our unified and dynamic graph (UniDG) is presented in Fig.\,\ref{fig:framework}. It consists of three main components including:
(1) A unified representation network to learn representations for different modalities within a single model.
(2) Dynamic graph clustering to construct more reliable affinity graphs via neighbors of varying quantities.
(3) Progressive 
Association method to fuse the multi-modal clustering results.
 

\subsection{Unified Representation Network}
%
%
%
Fig.\,\ref{fig:framework} manifests the basic structure of our unified representation network in compliance with Vision Transformer (ViT)~\cite{dosovitskiy2020image}. 
Given the face and body modalities in image conformation, we convert the audio-waveform voice into an image of a sequence of 128-dimensional log Mel-spectrogram. That means all modalities are firstly converted into the same image space. To balance the training and retain the modality uniqueness, we integrate all modalities into a unified label space, denoted as follows:
\begin{equation}  
	L = \{\underbrace{1,..,F}_{L^f};\underbrace{F+1,..,F+B}_{L^b};\underbrace{F+B+1,..,F+B+V}_{L^v}\}.
\end{equation}

The unified label space makes it possible to train different modalities in the same space. Such a label union also guarantees unique semantics in different modalities since all classes are preserved.
As a result, only one single classifier are necessitated for modalities, which contributes to stable training. 
Fig.\,\ref{fig:framework} shows that the network firstly splits an input samples $x_i\in \mathbb{R}^{h\times w}$ with label $y_i \in L$ into $N$ non-overlapping patches. The $j$-th patch is then linearly projected into a token $z^0_{i,j} \in \mathbb{R}^d$. All tokens are concatenated into a sequence $\mathbf{z}_i^0$ associated with a learnable classification token $z^0_{i, cls}$ as:
\begin{equation}
\mathbf{z}_i^0=[z^0_{i, cls}, z^0_{x_{i,1}}, z^0_{x_{i,2}},..., z^0_{x_{i,N}}] + \textbf{p},
\end{equation} 
where $\textbf{p}\in \mathbb{R}^{N\times d}$ denotes a learnable positional embedding.
%

%
The tokens are then passed through an encoder consisting of a sequence of $L$ Transformer layers. Each layer $l$ comprises of the multi-head attention ($MSA$)~\cite{vaswani2017attention}, layer normalization ($LN$)~\cite{ba2016layer}, and sequentially-stacked fully connected layers ($MLP$) as follows:
\begin{align}
\mathbf{t}_i^l&=MSA\big(LN(\mathbf{z}_i^{l-1})\big)+\mathbf{z}_i^{l-1},\\
\mathbf{z}_i^{l}&=MLP\big(LN(\mathbf{t}_i^l)\big)+\mathbf{t}_i^l.
\end{align}

Finally, the transformed class token $z^L_{i,cls}$ plays as feature representation of input $x_i$. It is sent to a classifier to derive its class probability vector $p_i$. To optimize the network, we leverage the cross-entropy loss and triplet loss as:
\begin{equation}
    \mathcal{L} = \mathcal{L}_{ce} + \mathcal{L}_{triplet},
\end{equation}
where $\mathcal{L}_{ce}=-log(p_{i,y_i})$ and $\mathcal{L}_{triplet} = max\big(0, m+D(z^L_{i,cls},z^L_{p,cls})-D(z^L_{i,cls},z^L_{n,cls})\big)$. Here, $D(\cdot,\cdot)$ return Euclidean distance of its two inputs and $m$ denotes a pre-defined margin. The $z^L_{p,cls}$/$z^L_{n,cls}$ denote sample representation of the same/different labels to $x_i$.




\subsection{Dynamic Graph Clustering}

After pre-training the unified network, the next comes to construct dynamic graph clustering based on the learned representations.


\begin{figure}[t]
	\centering
 	\includegraphics[width=\linewidth]{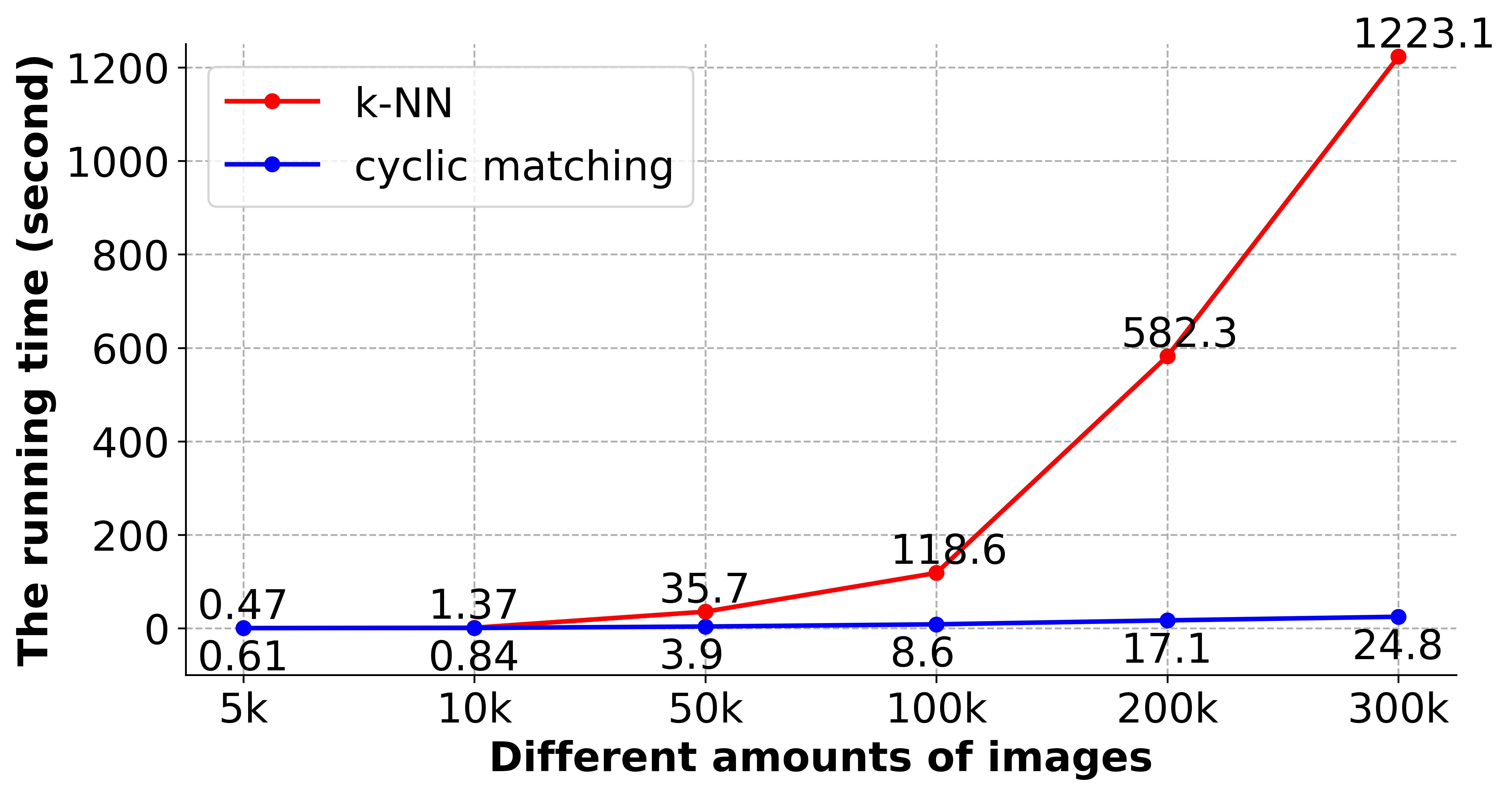}  
	\caption{Graph building time at different data sizes. The experiments are performed on a large-scale face dataset Web-Face42M~\cite{2021WebFace260M}.
	}
    \label{fig:running-time}
\end{figure}

%
\textbf{Cyclic Matching}.
Most GCN-based clustering methods ~\cite{wang2019linkage,yang2020learning,shen2021structure} utilize $k$-NN to construct affinity graphs. They consider a fixed $k$ value for all nodes, which however, is not always the case. The nearest neighbors are unique for each object and fixed neighbors may lead to non-negligible performance damages. 
%
%
To solve this issue, we propose to explore dynamic neighbors for each node by further trimming implausible connections from the vanilla $k$-NN.
%
%
Specifically, we adopt a simple-yet-strong cyclic matching strategy to update the neighbors for all nodes. This strategy is used as post processing in re-ID~\cite{zhong2017re}, but we use it in the training phase to obtain dynamic graph.

An intuitive example is manifested in Fig.\,\ref{fig:framework} where node ``c'' is the top-2 neighbor of node ``a'', but node ``a'' ranks top-10 of node ``c''. Assuming the ranking threshold is set to 3, node ``a'' and ``c'' should not be neighbors and the connection between them should be withdrawn.
Based on this connection removal strategy, we cyclically update each node to finally get a dynamic graph with variable neighbors.
Although this strategy involves a bit of computation, its overhead is trivial compared to the $k$-NN graph construction. As shown in Fig.\,\ref{fig:running-time}, we compare the time of graph construction at different data sizes. With the increase of data scale, the time required for \emph{k}-NN increases exponentially, but the cyclic matching takes negligible time overhead. As this work study the video character grouping, the shot-level frames (\textless 10k) in each video would be much smaller than public face datasets (\textgreater 1M). The cyclic matching strategy costs little time in this setting.

\begin{figure}[t]
	\centering 
	\includegraphics[width=\linewidth]{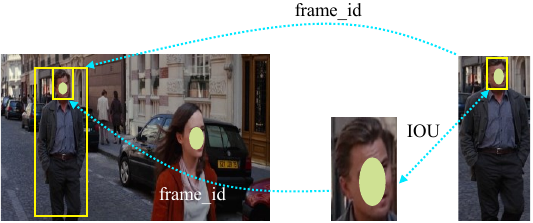} 
	\caption{Spatial association.
	The spatial IOU is leveraged to match the face and body. As the face and body belong to one specific frame, we need to match them in only that frame, and the matching is very fast.
	} 
	\label{fig:spatio}
\end{figure}   

\begin{figure}[t]
	\centering
	\includegraphics[width=\linewidth]{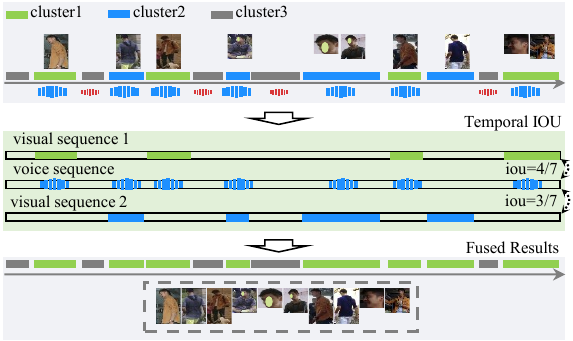}  
	\caption{Temporal association.
	The temporal IOU is leveraged to match the voice sequence and other visual sequences. If two visual sequences have high IOU values with the same voice sequence, they may belong to the same identity. 
	} 
	\label{fig:temporal}
\end{figure}


%
\textbf{Graph Clustering}.
We dive into a full description about our graph clustering for both supervised and unsupervised settings. As shown in Fig.\,\ref{fig:framework}, to preserve global instance-wise information along network training, we build a memory to cache the learned representations, which are extracted through our unified representation network. The cached representations are used for fully graph optimization.


For supervised clustering, state-of-the-art methods are based on graph convolutional networks. We firstly generate the affinity graph using \emph{k}-NN, then update each nodes with variable neighbors to obtain the dynamic graphs. Finally, the dynamic graphs are input to the GCN network to conduct clustering. 
For unsupervised clustering, we also apply same strategy to update the cached representations. 
Considering the unavailability of labels.
we 
refine the input representations in a weighted fashion. Supposing a total of $N$ representations are stored in the cache, each representation is updated with its cyclic neighbors as:
\begin{equation}  
    \widehat{f_{i}} = f_{i} + \sum_{j=1}^{n} (w_{i,j} \times f_{j}),
    \label{eq9}
\end{equation}
where $i \in \left[1,N\right]$. The $n$ denotes number of cyclic neighbors in the cache and it is variable for different nodes. The $w_{i,j}$ is the weighted parameter between $f_{i}$ and its $n$ cyclic neighbors and it is formulated as $w_{i,j} = e^{-d_{i,j}}/{\sum_{j=1}^{n} e^{-d_{i,j}}}$ where $d_{i,j}$ refers to the cosine distance between $f_{i}$ and $f_{j}$. 







\subsection{Progressive Association}
Aftering clustering all modalities, we need to fuse them and output all appearing moments of major characters. As shown in Fig.\,\ref{fig:framework}, we propose a progressive association method to perform the multi-modal fusion. It fully considers the spatiotemporal contexts among modalities. Both utilize the IOU value to conduct matching, which is defined as follows: 
\begin{equation}
    IOU = \frac{S^f\cap S^b}{S^f\cup S^b},
\end{equation}
where $S^f$ and $S^b$ indicate the spatial boxes or temporal sequences of the face and body modalities, respectively.



%
\textbf{Spatial Association}.
In our study, we find that face has higher precision than body and voice, but cannot recall some moments with invisible faces, \emph{e.g.,} far away from the camera or back to the camera. To this end, built upon the face clustering results, we gradually integrate the body clustering as assistance as shown in Fig.\,\ref{fig:spatio}. Specifically, we first associate the face with its corresponding body by calculating the spatial IOU. Assuming the coordinates (upper left and lower right corners) of the face are $(x_1^f, y_1^f)$ and $(x_2^f, y_2^f)$, and the coordinates of the body are $(x_1^b, y_1^b)$ and $(x_2^b, y_2^b)$, the spatial IOU value is defined as:
\begin{equation} 
\begin{split}
    \label{eq:iou_spatial}
    IOU_{spatial} = & \frac{[min(x_2^f, x_2^b)-max(x_1^f, x_1^b)]}{[max(x_2^f, x_2^b)-min(x_1^f, x_1^b)]} \cdot
    \\&
    \frac{[min(y_2^f, y_2^b)-max(y_1^f, y_1^b)]}{[max(y_2^f, y_2^b)-min(y_1^f, y_1^b)]}
\end{split}
\end{equation}

Since the face and body only belong to a specific frame, we only need to conduct a fast matching upon that frame. Then, the pseudo-labels of bodies are updated by the face labels, and other samples in the same cluster can be assigned with the same label. For those clusters that have two or more face labels, we select the label with the highest number of faces by voting. Based on the spatial association, the labels of the face and body samples are unified. In particular, those moments with invisible faces can be recalled.

\begin{figure*}[!t] 
	\centering
	\includegraphics[width=13cm]{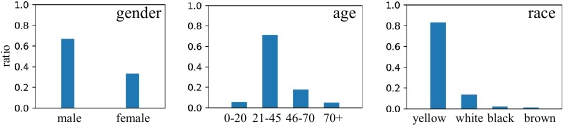} 
	\caption{Demographic distributions in the MTCG. This part of the statistical characteristics are obtained by manual statistics.
	} 
	\label{fig:distribution}
\end{figure*} 

\begin{table*}[t]
    \small
	\centering  
	\caption{Data statistics of VPCD and MTCG.
		{\#}EPS: number of programs; {\#}IDs: number of characters; RD: raw data.}
	\renewcommand\arraystretch{1.2} 
			\begin{tabular}{p{2cm}|p{1.2cm}<{\centering}p{0.8cm}<{\centering}p{0.8cm}<{\centering}p{0.8cm}<{\centering}|p{1.2cm}<{\centering}p{1.2cm}<{\centering}p{1.2cm}<{\centering}}
				\thickhline
				\textbf{Dataset} & \textbf{Length} & {\#}\textbf{EPS} & {\#}\textbf{IDs} & \textbf{RD} & {\#}\textbf{Face} & {\#}\textbf{Body} & {\#}\textbf{Voice} \\
				\hline 
				\hline
				VPCD~\cite{Brown2021FaceBV} & 23h54m & 6 & 326 && {35,396} & {39,777} & {9,165}  \\
		        \hline
				MTCG & {34h6m} & {21} & {361}& \checkmark & 26,584 & {41,770} & {8,558} \\
				\thickhline
				
		\end{tabular}
		\label{tab:dataset-static}

\end{table*}

\begin{figure*}[ht]
	\centering
	\includegraphics[width=15cm]{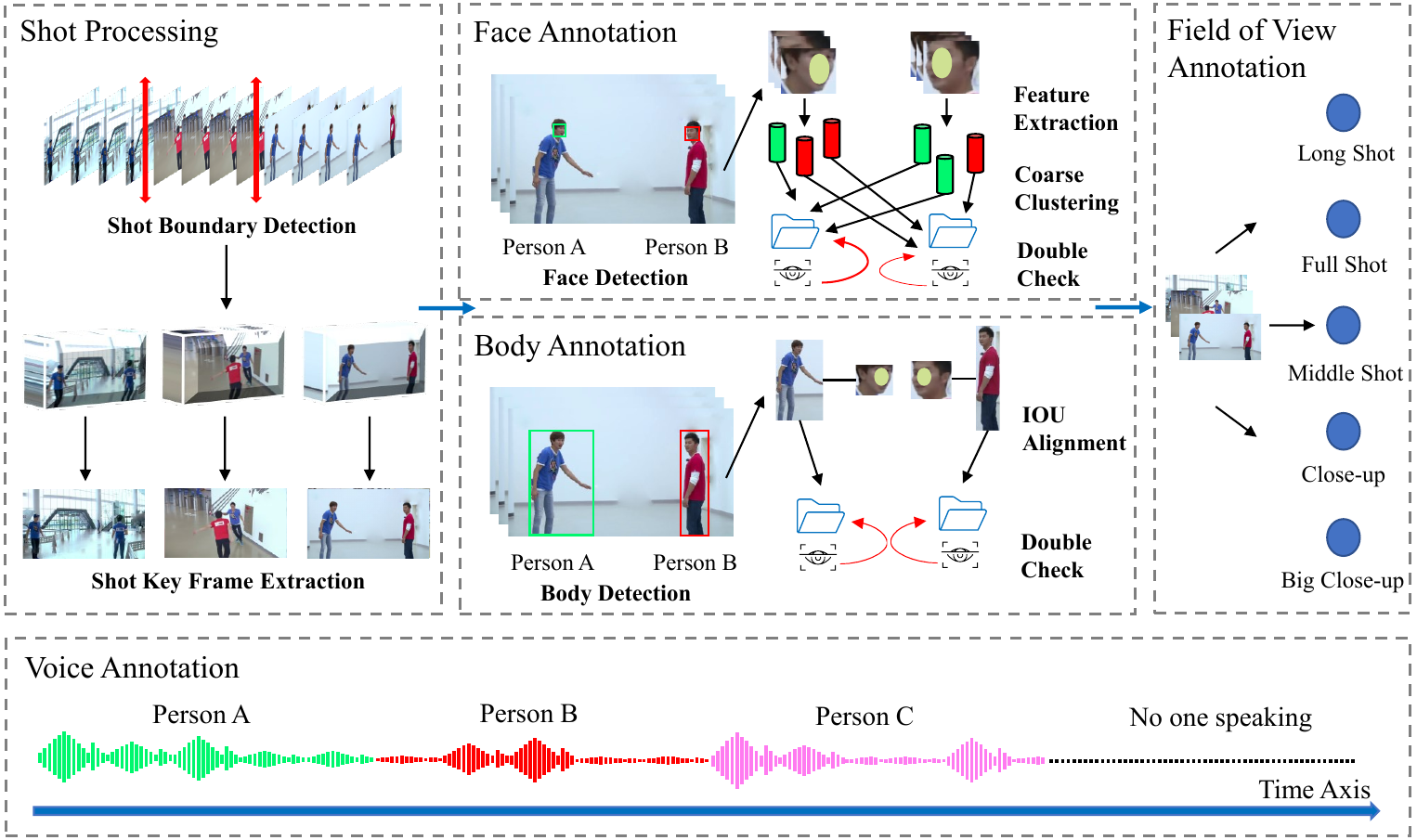} 
	\caption{The annotation pipeline of MTCG.  
	The annotation pipeline is divided into five parts --- shot processing, face annotation, body annotation, field of view annotation and voice annotation. 
	} 
	\label{fig:annotation_process}
\end{figure*} 

\begin{figure*}[t]
	\centering
	\includegraphics[width=0.9\linewidth]{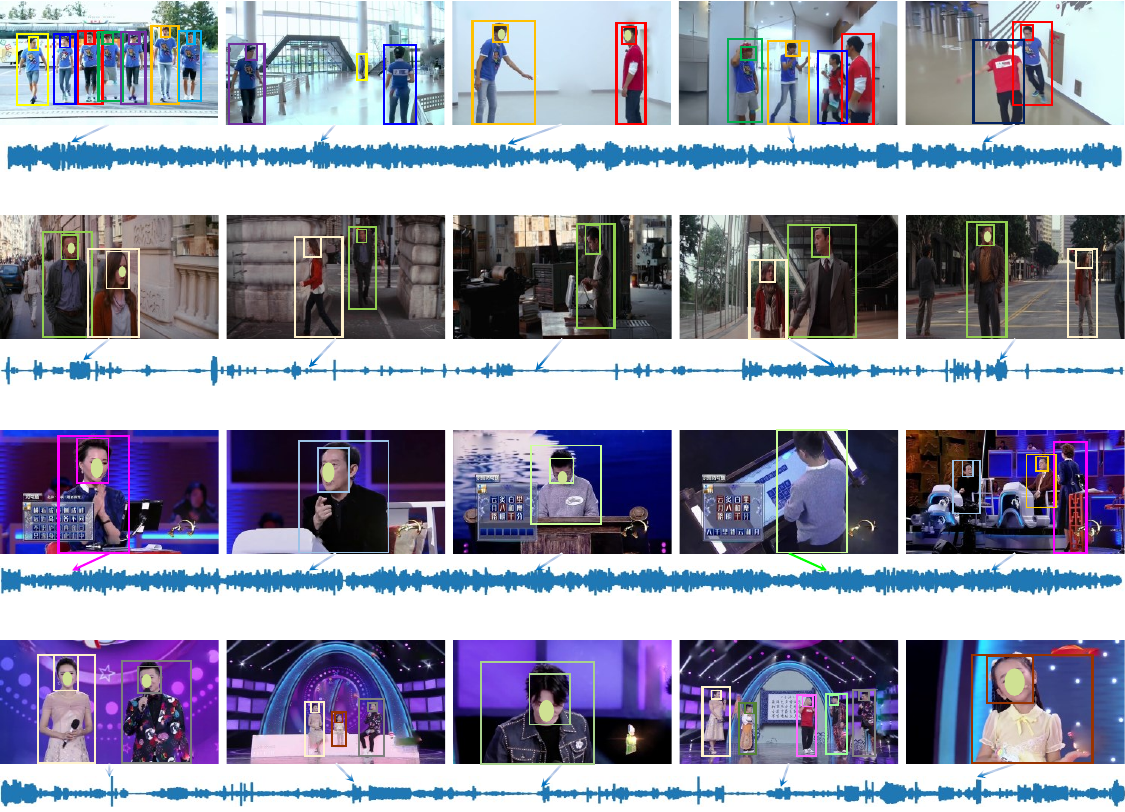}  
	\caption{Visualization of samples in MTCG. 
 The frames in each row are extracted from the same video. 
 Each frame represents a shot of the video. The bounding boxes of the same color denote person of the same identity.
	}
	\label{fig:sample}
\end{figure*}

%
\textbf{Temporal Association}.
After spatial association, some clusters with the same identity are still separated due to the existence of some hard samples with low quality or significant appearance changes. To solve this issue, we leverage the voice tracks to bridge separate clusters. Specifically, in movies or TV shows, characters are usually accompanied by dialogues, so the voice has correlations with faces or bodies in time series. A cluster of voices records a person's speaking trajectory and a cluster of faces/bodies also records the visual trajectory. By calculating the temporal IOU, two separate clusters can be fused if they both have high IOU values with the same voice trajectory. Assuming the visual and voice sequences are denotes as $\{(s_i^{fb},e_i^{fb})\}|_i^M$ and $\{(s_j^{v},e_j^{v})\}|_j^N$, where $s_i$ and $e_i$ are the start and end times of the $i^{th}$ continuous sequence of shots. $M$ and $N$ denote the total number. $fb$ denotes the fused face and body clusters. $v$ denotes the voice clusters. The temporal IOU is defined as:
\begin{equation}
    \label{eq:iou_temporal} 
    IOU_{temporal} = \sum_i\sum_j\frac{|min(e_i^{fb}, e_j^v)-max(s_i^{fb}, s_j^v)|}{|max(e_i^{fb}, e_j^v)-min(s_i^{fb}, s_j^v)|}.
\end{equation}

Fig.\,\ref{fig:temporal} gives a toy example. The man wears completely different clothes on different occasions, leading to isolated clusters (\textcolor{green}{green} and \textcolor{blue}{blue}) even fusing face and body clusters. Since the face and body images in each cluster record their frame numbers, each cluster can be regarded as a visual sequence in the temporal dimension. If two visual sequences both have high temporal IOU values with the same voice sequence, they may have the same identity. This is reasonable since visual and voice sequences should be highly correlated if the same person appears in multiple locations.

To avoid false associations, we only select the voice trajectories with high confidence. This method will reduce the number of clusters by a certain amount, making the clusters more compact. After fusion, each cluster denotes a character's appearing moments, thus all major characters are obtained. Note our fusion method is different from VPCD~\cite{Brown2021FaceBV}. VPCD only clusters faces and utilizes a speaking track to merge face clusters, requiring two clusters to be close in time. Our fusion clusters three modalities and considers the whole trajectory, enabling to merge of two clusters with a long time interval.

\section{The MTCG Dataset}
To evaluate our method, we collect a multi-modal dataset with shot-level annotations. The collected video dataset is termed Multi-modal Temporal Character Grouping (MTCG). This section gives a concise description of the statistical data and annotations of this dataset, refer to the supplementary materials for a more detailed description.


\subsection{Statistical Data}
Different from VPCD~\cite{Brown2021FaceBV} that only provides pre-extracted features, MTCG provides each character’s raw images, face/body, and speaking voice clips. As shown in Table\,\ref{tab:dataset-static}, MTCG covers around 34-hour-long videos consisting of 21 different TV shows and movies. It contains 26,584 face images, 41,770 body images, and 8,558 voice clips, respectively. Fig.\,\ref{fig:distribution} gives the statistical distribution of gender, age, and race of all characters, and Fig.\,\ref{fig:sample} visualizes some samples in MTCG. Based on MTCG, researchers can explore more effective representations for multi-modal temporal character grouping.

 
\subsection{Annotation Process}
Fig.\,\ref{fig:annotation_process} gives an illustration of the annotation pipeline. Here, we give a more detailed description about the annotation for MTCG. Note that all the annotations were done by professional annotators.


\textbf{Shot processing}.
Shot is the basic unit of video. As the variation within each shot is very little, we extract only one frame per shot to avoid information redundancy. In particular, we use TransNet~\cite{tu2017transnet} as a shot boundary detector to split a whole video into several consecutive shots. The middle frame of each shot is retained.

\textbf{Face annotation}.
We first use the trained detector to detect faces, then extract the face features and cluster them into some coarse groups with traditional clustering algorithms. 
Next, we manually combine the face groups with the same identity and correct the wrong samples in each face group. In this way, we can quickly and accurately classify the main characters in a video.

\textbf{Body annotation}.
Similar to the face annotation, high-quality body bounding boxes are obtained through a pre-trained detector. We first align each body with its face by computing the IOU value, then obtain its character identity. Next, we double-check the alignment results and correct the mistaken ones. For bodies that could not be aligned to any faces, we manually assign them to corresponding clusters.



\textbf{Voice annotation}. 
We first leverage the tool \emph{ffmpeg} to extract the voice signals, then use ASR to detect the periods where somebody is speaking. Finally, we manually align a person identity for each period and check the speaking periods to ensure reliability. 

\textbf{Post Processing}.
All the annotated results are aggregated to get all the moments of major appearing characters. MTCG gives abundant annotations of three modalities. For instance, each visible face/body has the corresponding bounding box, confidence and shot type. For all samples, we record their spatial and temporal locations in the video. Aligned face images are also provided. All voice tracks are manually labeled to ensure reliability.

\section{Experiments}

In this section, we evaluate our method through three sets of experiments: face clustering, multi-modal character grouping, and retrieval experiments. 




\begin{table}[t]
	\centering 
		\caption{Comparison of face-clustering on \text{MTCG}. All the methods except the last one utilize the same representation extracted by VGGFace2. Our baseline method is L-GCN. U: unsupervised; SS: self-supervised; S: supervised. ``HAC'' : Agglomerative Clustering. 
        The last experiment utilize the representation extracted by our URN. Noted that VGGFace2 method and our URN are pre-trained on the same VGGFace2 dataset.
		}
	\small
	\renewcommand\arraystretch{1.2}
			\begin{tabular}{p{3.1cm}|p{0.7cm}<{\centering}|p{0.8cm}<{\centering}p{0.8cm}<{\centering}p{0.8cm}<{\centering}} 
				\thickhline
				\textbf{Method} &  \textbf{Type} & $F_P$ & $F_B$ &  $NMI$ \\
				\hline
				\hline
				$k$-Means~\cite{lloyd1982least}& \multirow{7}{*}{U}& 31.2 & 35.4 & 40.0 \\
				DBSCAN~\cite{ester1996density}&  & 35.9 & 46.8 & 40.4 \\
				Spectral Clustering~\cite{fowlkes2004spectral}&  & 31.5 & 34.7 & 38.6 \\
				HAC~\cite{murtagh2014ward}&  & 34.1 & 38.3 & 43.2 \\
				HDBSCAN~\cite{mcinnes2017hdbscan}&  & 31.6 & 47.3 & 33.7 \\
                $k$-sums~\cite{pei2020efficient}&  & 28.6& 35.2 & 40.5 \\
                FINCH~\cite{sarfraz2019efficient}&  & 27.7&35.6 & 47.1  \\
                \hline
                SSiam~\cite{sharma2019self}& \multirow{2}{*}{SS} &33.0 & 36.8 & 41.5 \\
                FFG~\cite{rothlingshofer2019self}&  & 35.4& 37.9 & 46.6  \\
		        \hline
				GCN-D~\cite{yang2019learning}&\multirow{4}{*}{S}  & 19.4 & 28.5 & 53.2 \\
				L-GCN~\cite{wang2019linkage}&  & 41.9 & 51.4 & 52.3 \\
				GCN-V~\cite{yang2020learning}&  & 42.6 & 44.8 & 44.8 \\
				STAR-FC~\cite{shen2021structure}&  & 35.9 & 47.8 & 29.6 \\
				\hline
				Ours (face, DGC)& \multirow{2}{*}{S} & 46.3 & 52.2 & 56.4 \\
				Ours (face, URN, DGC)&  & \textbf{77.2} & \textbf{83.6} & \textbf{82.5} \\
	            \thickhline
				
		\end{tabular}
		\label{tab:cluster-baseline}
\end{table}

\begin{table}[!t]
		\caption{Face, body and voice clustering on VPCD. ``DGC'' denotes the dynamic graph clustering. Bold numbers mean ``+DGC'' brings gains. The table reports averaged metrics across all six program sets. Unsupervised $k$-Means and Hierarchical Agglomerative Clustering are tested on complete set of VPCD, while supervised L-GCN is firstly trained on train set (80$\%$ data of VPCD) and then tested on test set (20$\%$ data of VPCD).
		}
	\centering 
	\small
	\renewcommand\arraystretch{1.2} 
			\begin{tabular}{p{1.8cm}p{0.6cm}<{\centering}|p{0.3cm}<{\centering}p{0.3cm}<{\centering}p{0.6cm}<{\centering}|p{0.3cm}<{\centering}p{0.3cm}<{\centering}p{0.6cm}<{\centering}}
				\thickhline
				\multirow{2}{*}{\textbf{Method}} &  \multirow{2}{*}{\textbf{Mode}} & \multicolumn{3}{c|}{\textbf{baseline}} & \multicolumn{3}{c}{\textbf{baseline+DGC}}\\
				
				\cline{3-8}    
				 &  &  $F_P$ & $F_B$ & $NMI$ & $F_P$ & $F_B$ & $NMI$ \\
				\hline 
				\hline
				$k$-Means~\cite{lloyd1982least} & & 35.1 & 44.3 & 63.8 & \textbf{39.4} & \textbf{47.1} & \textbf{65.6}  \\
				HAC~\cite{murtagh2014ward} & Face & 40.3 & 47.6 & 65.4 & \textbf{42.0} & \textbf{49.0} & \textbf{66.1} \\
				L-GCN~\cite{wang2019linkage} & & 48.1 & 50.6 & 51.3 & \textbf{51.9} & \textbf{56.5} & \textbf{55.6}  \\
				\hline
				$k$-Means~\cite{lloyd1982least} & & 22.7 & 30.4 & 47.5 & \textbf{23.2} & \textbf{31.4} & \textbf{48.7}  \\
				HAC~\cite{murtagh2014ward}  & Body & 23.5 & 32.2 & 49.1 & \textbf{24.3} & \textbf{32.7} & \textbf{49.9} \\
				L-GCN~\cite{wang2019linkage} & & 27.3 & 30.9 & 43.6 & \textbf{30.7} & \textbf{37.6} & 42.3  \\
				\hline
			    $k$-Means~\cite{lloyd1982least} & & 25.7 & 32.6 & 49.2 & \textbf{27.9} & \textbf{35.4} & \textbf{50.3} \\
				HAC~\cite{murtagh2014ward}  & Voice & 25.9 & 32.0 & 48.2 & \textbf{27.9} & \textbf{34.4} & \textbf{49.5} \\
				L-GCN~\cite{wang2019linkage} & & 30.2 & 33.1 & 24.6 & \textbf{30.5} & \textbf{38.1} & 21.5  \\
				\thickhline
				
		\end{tabular}
		\label{tab:VPCD}
	
\end{table}

\subsection{Implementation Details}

\textbf{Datasets}. The unified representation network is pre-trained on three large-scale datasets, Last~\cite{shu2021large} for body, VGGFace2~\cite{cao2018vggface2} for face, and Voxceleb2~\cite{Chung2018VoxCeleb2DS} for voice. For fair comparison, the compared methods (expert model in specific modality) are also pre-trained on them. For example, the face-modal methods are pre-trained on VGGFace2 and the body-model methods are pre-trained on LaST. The only difference is that we pre-trained our model on three modalities simultaneously. During the testing phase, the clustering evaluations are conducted on MTCG (16 videos for training and 5 videos for testing) and another three public datasets, $i.e.,$ 
VPCD~\cite{Brown2021FaceBV}, CASIA~\cite{2014Learning}(training), and  IJB-B~\cite{2017IARPA}(testing). The retrieval task is used to evaluate the representation on MTCG, MegaFace~\cite{kemelmacher2016megaface}, Voxceleb1~\cite{Nagrani17}, and Market-1501~\cite{zheng2015scalable}.

\begin{table}[!t] 
	\centering 
	\caption{Temporal character grouping on MTCG. $k$-Means and STAR-FC stand for unsupervised and supervised methods, respectively. ``*'' indicates evaluation in the shot-level aligned label space. All the features in this table are extracted by our URN. 
 } 
	\renewcommand\arraystretch{1.2} 
	\small
	\begin{tabular}{p{1.6cm}|p{0.6cm}<{\centering}p{0.6cm}<{\centering}p{0.7cm}<{\centering}|p{0.6cm}<{\centering}p{0.6cm}<{\centering}p{0.7cm}<{\centering}}  
		\thickhline
		\multirow{2}{*}{\textbf{Methods}} &  \multicolumn{3}{c|}{\textbf{$k$-Means~\cite{lloyd1982least}}} & \multicolumn{3}{c}{\textbf{STAR-FC~\cite{shen2021structure}}} \\ 
		
		\cline{2-7} 
		& \textbf{$F_P^*$}  & \textbf{$F_B^*$}    & \textbf{$NMI^*$}
		& \textbf{$F_P^*$}  & \textbf{$F_B^*$}    & \textbf{$NMI^*$}\\ 
		\hline		 
		\hline		 
		Face   
		&39.9 &52.4 &53.9 &48.2 &58.1 &58.7 \\ 
		
		Body  
		&40.2 &48.9 &54.6 & 42.8  & 52.0   & 57.7\\ 
		
		Voice  
		&16.9 &23.2 &25.1 & 10.8  & 20.7  & 44.5\\
		\hline
		Fusion  
		&\textbf{50.5} &\textbf{57.4} &\textbf{60.7} &\textbf{55.1} &\textbf{61.8} &\textbf{63.5}\\
		
		\thickhline
	
	\end{tabular}
	\label{tab:multi_model_us} 
\end{table}

\textbf{Experimental Settings}. For data prepossessing, all samples are resized to 256$\times$128. For the unified representation network, we employ the widely-used ViT-B/16 as backbone, and SGD optimizer is employed with a momentum of 0.9 and the weight decay of 1e-4. The learning rate is initialized
as 0.008 with cosine learning rate decay.
To construct the dynamic graph, we set the K values of \emph{k}-NN as K=20 for MTCG, K=80 for VPCD, CASIA and IJB-B unless specified. We compared different supervised, self-supervised and supervised methods. Corresponding parameters follow the recommended values (by the respective authors) to report their best metrics scores.

\textbf{Metrics}.
We evaluate the performance on two tasks: clustering and retrieval. Clustering is measured by Pairwise F-score ($F_P$)~\cite{2009A}, BCubed F-score ($F_B$), and normalized mutual information ($NMI$)~\cite{2008Introduction}. For the character grouping, the metrics of three modalities are unified to shot level, and denoted as \textbf{$F_P^*$}, \textbf{$F_B^*$}, and \textbf{$NMI^*$}. Retrieval is measured by Cumulative Matching Characteristic (CMC) and Mean Average Precision (mAP). Following standard settings, we report top-1 (R1), top-5 (R5), and top-10 (R10) in the CMC curve. For ease of the following representation, we abbreviate our unified representation network as ``URN'' and dynamic graph clustering as ``DGC''.






\begin{figure*}
	\centering
	\begin{minipage}[t]{0.62\linewidth}
		\centering
        \vspace{0pt}
	\includegraphics[width=11.5cm]{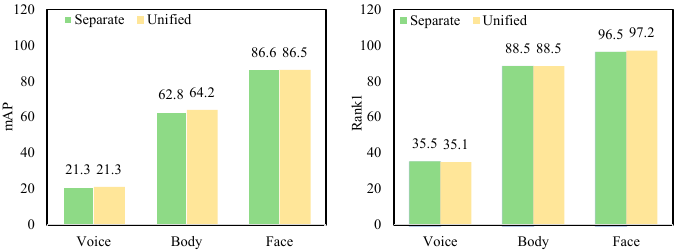}  
	\caption{Comparison of URN on separate training and unified training on MTCG. The separate method denotes the results of training with the corresponding modal data using the same architecture with URN. The results demonstrate that joint training with other modalities would not degrade the representation performance. 
	} 
	\label{fig:ablation}
	\end{minipage} 
	\quad
	\begin{minipage}[t]{0.32\linewidth}
		\centering   
        \vspace{0pt}
	\includegraphics[width=5.6cm]{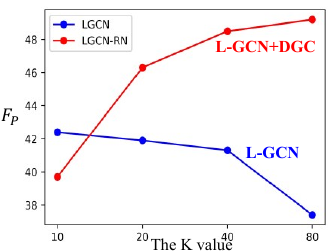}    
	\caption{The effect of the K value for dynamic graph clustering. Larger K value brings more precise neighbors, resulting better performance.
	} 
	\label{fig:kvalue}
		\end{minipage}  
\end{figure*}

\begin{table}[t]
	\centering
		\caption{Dynamic graph clustering with different current methods. ``*'' refers to using our dynamic graph clustering. 
		}%
	\small
	\begin{center}
		\renewcommand\arraystretch{1.2}  
		\begin{tabular}{p{2.1cm}|p{0.5cm}<{\centering}p{0.5cm}<{\centering}p{0.6cm}<{\centering}|p{0.5cm}<{\centering}p{0.5cm}<{\centering}p{0.6cm}<{\centering}}  
			    \thickhline
			    \multirow{2}{*}{\textbf{Dataset}} &  \multicolumn{3}{c|}{\textbf{MTCG\_face}} & \multicolumn{3}{c}{\textbf{CASIA\_IJB-B}} \\ 
		
		        \cline{2-7}    
				  &  $F_P$ & $F_B$ & $NMI$ & $F_P$ & $F_B$ & $NMI$ \\
				\hline 
				\hline

				$k$-Means~\cite{lloyd1982least} & 31.2 & 35.4 & 40.0 & 41.3 & 61.3 & 85.9  \\ 
				$k$-Means$^*$~\cite{lloyd1982least} & 33.3 & 37.4 & 41.9 & 44.1 & 63.6 & 86.8   \\ 
				\hline

				DBSCAN~\cite{ester1996density} & 35.9 & 46.8 & 40.4 & 3.0 & 56.1 & 56.8  \\ 
				DBSCAN$^*$~\cite{ester1996density} & 38.4 & 41.7 & 23.9 & 15.7 & 70.3 & 79.2  \\ 
				\hline

				L-GCN~\cite{wang2019linkage} & 41.9 & 51.4 & 52.3 & 51.3 & 74.8 & 88.2   \\ 
				L-GCN$^*$~\cite{wang2019linkage} & 46.3 & 52.2 & 56.4 & 54.3 & 75.9 & 89.4 \\ 
				\hline

				STAR-FC~\cite{shen2021structure} & 35.9 & 47.8 & 29.6 & 47.5 & 50.0 & 79.2    \\ 
				STAR-FC$^*$~\cite{shen2021structure} & 44.8 & 50.1 & 54.8 & 55.4 & 62.9 & 85.4    \\ 
				\thickhline
				
		\end{tabular}
		\label{tab:cluster-compare}
	\end{center}
\end{table}

\begin{table}[!t] 
	\centering 
	\caption{The effectiveness of Progressive Association on VPCD. ``F,B,V'' are short of face, body, and voice. ``*'' indicates evaluation in the unified label space. 
	} 
	\renewcommand\arraystretch{1.2} 
	\small
	\begin{tabular}{p{1.3cm}|p{0.5cm}<{\centering}p{0.6cm}<{\centering}p{0.7cm}<{\centering}|p{0.5cm}<{\centering}p{0.6cm}<{\centering}p{0.7cm}<{\centering}}  
		\thickhline
		\multirow{2}{*}{\textbf{Methods}} &  \multicolumn{3}{c|}{\textbf{k-Means~\cite{lloyd1982least}}} & \multicolumn{3}{c}{\textbf{Spectral Clustering~\cite{fowlkes2004spectral}}} \\ 
		
		\cline{2-7} 
		& \textbf{$F_P^*$}  & \textbf{$F_B^*$}    & \textbf{$NMI^*$}
		& \textbf{$F_P^*$}  & \textbf{$F_B^*$}    & \textbf{$NMI^*$}\\ 
		\hline		 
		\hline		 
		F   
		&53.9 &56.2 &54.2 &61.7 &61.8 &57.0 \\ 
		
		\hline
		F + B + V
		&\textbf{61.8} &\textbf{64.0} &\textbf{64.3} &\textbf{66.7} &\textbf{67.8} &\textbf{64.7}\\
		
		\thickhline
	
	\end{tabular}
	\label{tab:multi_model_VPCD} 
\end{table}

\begin{table}[!t]
		\caption{Fusion of stage1+HAC+stage3 on MTCG.
		}
	\centering 
	\small
	\renewcommand\arraystretch{1.2} 
			\begin{tabular}{p{0.3cm}<{\centering}p{0.3cm}<{\centering}p{0.8cm}<{\centering}|p{0.3cm}<{\centering}p{0.3cm}<{\centering}p{0.8cm}<{\centering} | p{0.3cm}<{\centering}p{0.3cm}<{\centering}p{0.6cm}<{\centering}}
			\thickhline
				\multicolumn{3}{c|}{Face} &  \multicolumn{3}{c|}{Fusion(HAC)} & \multicolumn{3}{c}{Fusion(HAC+DGC)}\\
				\hline 
				\hline 
\textbf{$F_P^*$}  & \textbf{$F_B^*$}    & \textbf{$NMI^*$} & \textbf{$F_P^*$}  & \textbf{$F_B^*$}    & \textbf{$NMI^*$} & \textbf{$F_P^*$}  & \textbf{$F_B^*$}    & \textbf{$NMI^*$} \\
                    \hline
                    41.9 & 55.3 & 60.5 & 51.0 & 61.3 & 69.9 & 54.0 & 63.2 & 70.3 \\

				\thickhline
				
		\end{tabular}
		\label{tab:HAC_fusion}
\end{table}

\subsection{Main results}\label{main_result}
\textbf{Face Clustering on MTCG}.
Since most works focus on face clustering, we first compare with them in the singe-modal setting. As shown in Table\,\ref{tab:cluster-baseline}, current methods can be divided into unsupervised, self-supervised, and supervised ones. For a fair comparison, all the experiments except the last one utilize the same representations extracted by VGGFace2. In this section, we leverage L-GCN as our baseline method. Table\,\ref{tab:cluster-baseline} shows us that our DGC achieves 46.3\%, 52.2\%, and 56.4\% on $F_P$, $F_B$, and $NMI$, behaving much better than other methods. Specifically, unsupervised methods usually rely on specific data assumptions, and self-supervised methods bring limited information, leading to their poor performance.
In general, supervised methods yield better results as they learn to capture different patterns according to various data distributions. Even so, our DGC outperforms the baseline method L-GCN by 4.4\%, 0.8\%, and 4.1\% on three metrics, demonstrating its effectiveness.

\textbf{Unified Representation for Clustering}.
Our MTCG dataset provides raw images and voices, making it possible to explore more powerful representations for clustering. 
In the last row of Table\,\ref{tab:cluster-baseline}, we conduct face clustering with the features extracted by our URN network. We can see that our unified representation brings impressive improvements, achieving 77.2\%, 83.6\%, and 82.5\%, respectively. The results have been improved by a large margin and are much better than the representations extracted by VGGFace2. This justifies the great potential of our unified representation for clustering.



\textbf{DGC applied to different modalities.}
Here we verify the effectiveness of dynamic graph clustering on VPCD pre-extracted features. Table\,\ref{tab:VPCD} reports face, body and voice clustering results. For most metrics, noticeable performance raise can be observed across all the face/body/voice modals when applying dynamic graph clustering on baselines, whether unsupervised or supervised ones. Our method enables neighboring samples to pull closer to the potential cluster centers in the early stage of clustering, thus forming more compact and accurate clustering results.


				


\subsection{Temporal Character Grouping}
Table\,\ref{tab:multi_model_us} compares the single-modal results and the multi-modal fusion ones. For three modalities, we utilize our URN to extract unified representations. $k$-Means directly clusters the features of the testing set, while STAR-FC is firstly trained on the training set in MTCG, then evaluated on the testing set. Table\,\ref{tab:multi_model_us} shows the multi-modal fusion results are much better than any single-modality results. The experiments verify the effectiveness of our method for multi-modal temporal character grouping.



\begin{table}[t]
		\caption{Retrieval comparison on MTCG.``F'', ``V'', ``B'' are short of face, voice, and body, respectively. All the SOTA methods have been pre-trained with corresponding data, \emph{i.e.,} VGGFace2 for face, Last for body, Voxceleb2 for voice, and our URN uses them all for joint pre-training then is directly tested on MTCG.
		}
	\centering 
	\small
	\renewcommand\arraystretch{1.2} 
			\begin{tabular}{p{1.45cm}p{0.5cm}<{\centering}p{1.1cm}<{\centering} |p{0.6cm}<{\centering}p{0.55cm}<{\centering}p{0.55cm}<{\centering}p{0.55cm}<{\centering}}
			\thickhline
				\textbf{Method} &  \textbf{Mode} & \textbf{Dataset} &  \textbf{mAP} & \textbf{R1} & \textbf{R5} & \textbf{R10}\\
				\hline 
				\hline 
				CosFace~\cite{wang2018cosface}& \multirow{4}{*}{Face} & \multirow{4}{*}{MTCG-F} & 75.4 & 94.7 & 97.2 & 98.4 \\
		
				VPL~\cite{deng2021variational} &  &  & 78.2 & 95.6 & 98.7 & 99.1 \\
				
				ElasticFace~\cite{boutros2021elasticface} &  &  & 77.2 & 96.2 & 99.1 & 99.4 \\
				Ours &  &   & \textbf{86.5} & \textbf{97.2} & \textbf{99.4} & \textbf{99.7} \\ 
				\hline
				%
				TransReid~\cite{He_2021_ICCV}  &\multirow{3}{*}{Body} & \multirow{3}{*}{MTCG-B}& 62.8 & \textbf{88.5} & 94.5 & 96 \\
				BoT~\cite{luo2019bag}  &  &  & 51.1 & 82.5 & 92 & 94.5\\

				Ours  &  &  & \textbf{64.2} & \textbf{88.5} & \textbf{95.7} & \textbf{96.8} \\
				\hline
				IDM~\cite{chung2020in} & \multirow{4}{*}{Voice} & \multirow{4}{*}{MTCG-V}  & 15.3 & 30.1 & 47.6 & 56.4\\
				
				PSLA~\cite{gong2021psla}  &  &   & 14.3 & 31.1 & 49.3 & 61.8\\
				%
				%
				
				Titanet~\cite{koluguri2021titanet}  &  &   & 16.3 & 31.4 & 54.7 & 67.2 \\
				
				%
				Ours  &  &  & \textbf{21.3} & \textbf{35.1} & \textbf{60.8} & \textbf{73.3}\\
				\thickhline
				
		\end{tabular}
		\label{tab:repre-res-fbv}
\end{table}

\begin{table}[t]
	\centering 
		\caption{Retrieval Comparison on public datasets. ``MF'', ``MK'', ``Vox1'' denote MageFace, Market-1501, and Voxceleb1. Pre-training settings are the same as Table\,\ref{tab:repre-res-fbv}.
		}
	\small
	\renewcommand\arraystretch{1.2} 
			\begin{tabular}{p{1.45cm}p{0.5cm}<{\centering}p{0.9cm}<{\centering} |p{0.6cm}<{\centering}p{0.6cm}<{\centering}p{0.6cm}<{\centering}p{0.6cm}<{\centering}}
				\thickhline
				\textbf{Method} &  \textbf{Mode} & \textbf{Dataset} &  \textbf{mAP} & \textbf{R1} & \textbf{R5} & \textbf{R10}\\
				\hline 
				\hline
				CosFace~\cite{wang2018cosface} & \multirow{4}{*}{Face} & \multirow{4}{*}{MF} & 80.5 & 95.6 & 97.8 & 98 \\

				VPL~\cite{deng2021variational} &  &  & 77.4 & 94.3 & 97.1 & 97.7 \\
				
				ElasticFace~\cite{boutros2021elasticface} &  &  & 83.3 & 96.3 & \textbf{98.5} & \textbf{98.6} \\
				
				Ours &  &  & \textbf{84.5} & \textbf{97.3} & 98.4 & \textbf{98.6} \\
				\hline
				TransReid~\cite{He_2021_ICCV}  &\multirow{3}{*}{Body} &\multirow{3}{*}{MK} & 24 & 49.2 & 65.4 & 71.9\\
				BoT~\cite{luo2019bag} &  &  & 11.2 & 29 & 45.2 & 52.7\\

				Ours  &  &  & \textbf{26.1} & \textbf{50.8} & \textbf{67.5} & \textbf{74.9}\\
				\hline
			    IDM~\cite{chung2020in}  &\multirow{4}{*}{Voice} &\multirow{4}{*}{Vox1}  & 80.7 & 98.7 & \textbf{100} & \textbf{100}\\
				
				PSLA~\cite{gong2021psla}  &  &  & 63.5 & 98.5 & 99.6 & 99.8\\
				%
				%
				Titanet~\cite{koluguri2021titanet}  & &  & 54.3 & 94.5 & 99.1 & 99.4 \\
				Ours  &  &  & \textbf{81.4} & \textbf{99.8} & \textbf{100} & \textbf{100}\\
				\thickhline
				
		\end{tabular}
		\label{tab:repre-res-public}
	
\end{table}


\subsection{Ablation Study}\label{ablation-sec}

\textbf{Unified Training \& Separate Training}.
For fair comparison, we leverage the same architecture for training on MegaFace, Voxceleb1, and Market-1501, respectively. Then, the three pre-trained models are compared with our unified model. As shown in Fig.\,\ref{fig:ablation}, the unified-trained URN achieves competitive results compared with separate-trained models, and even slightly better in some cases. For example, the unified URN achieves a 1.4\% boost in mAP for the body modality, and +0.7\% in Rank-1 for the face modality. This shows unified representation be competitive compared with expert models.


\textbf{Effectiveness of the DGC}.
To further demonstrate the effectiveness of the dynamic graph clustering, we re-train and test several methods on MTCG and another two face datasets, CASIA and IJB-B. Here the features of MTCG come from VGGFace2, and the features of CASIA and IJB-B are from ArcFace~\cite{deng2019arcface}. Table\,\ref{tab:cluster-compare} shows: (1) For unsupervised methods, dynamic graph clustering brings a little gains in most metrics. (2) For supervised methods, our dynamic graph clustering has achieved a significant performance improvement. 
These experiments reveal huge application potential of the dynamic graph clustering in both unsupervised and supervised clusterings.

\textbf{Effects of the K Value on Different methods}.
Our main hyper-parameter in clustering is the K value. It can be set based on the number of samples per identity. Usually the larger the number of samples, the larger the value of K. As shown in Fig.\,\ref{fig:kvalue}, only when K is small, L-GCN is better than L-GCN+DGC. When K is large, such as K=80, $F_P$ of L-GCN reaches 37.4\%, but that of L-GCN+DGC reaches 49.2\%. With DGC, larger K brings better results, this is because more positive neighbors are connected. Conversely, larger K brings more false edges when constructing graphs using fixed-neighbor k-NN. The results show that our dynamic clustering is robust enough under different settings.


\begin{figure}[t]
	\centering
 	\includegraphics[width=\linewidth]{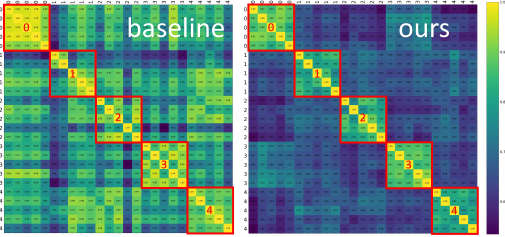}  
	\caption{Comparison of feature similarities between the VGGFace2 baseline and our URN method. 
	The bright yellow region means two features are similar, while the dark blue one means dissimilar. The red boxes cover all the samples in the same clusters.
	}
	\label{fig:similarity}
\end{figure}

\subsection{Multi-modal Fusion on VPCD}
In this section, we evaluate our Progressive Spatio-Temporal Association (PSTA) strategy on the VPCD~\cite{Brown2021FaceBV} dataset. The experimental results are shown in Table\,\ref{tab:multi_model_VPCD}. Since the authors of VPCD have not released the code, their method to compute the multi-modal metrics after fusion is unknown. In our experiments, all the positive shots in which a person appears would be taken into account when computing the multi-modal or uni-modal metrics (marked with ``*'').  Table\,\ref{tab:multi_model_VPCD} shows us that both k-Means~\cite{lloyd1982least} and Spectral Clustering~\cite{fowlkes2004spectral} achieve a large margin of improvement after the multi-modal fusion. The results further demonstrate the effectiveness of our PSTA strategy.

\subsection{Previous Methods inserted in our Pipeline}
Previous experiments have verified the effectiveness of our unified representation and dynamic clustering. We are still curious about what happens when the previous method is inserted into our pipeline. Here, we leverage HAC(ward) to replace our clustering and evaluate it on the MTCG. As shown in Tabel\,\ref{tab:HAC_fusion}, a significant performance improvement (from 41.9\% to 51.0\% on $F_P$) has been seen from single-modal to multi-modal results. Also, we can see that our DGC brings impressive improvements, achieving from 51.0\% to 54.0\% on $F_P$. This justifies the effectiveness of our dynamic clustering and fusion strategy.


\subsection{Retrieval Experiments}
In this section, we leverage retrieval experiments to further demonstrate the effectiveness of our unified representation.

\textbf{Retrieval on MTCG}.
We first evaluate our unified representation network (URN) on MTCG. 
As shown in Table\,\ref{tab:repre-res-fbv}, our URN outperforms other methods across all modalities. In the face mode, we compare with margin-based methods CosFace~\cite{wang2018cosface}, ElasticFace~\cite{boutros2021elasticface}, and prototype-based method VPL~\cite{deng2021variational}. Compared with ElasticFace, we achieve a performance gain of 9.3\% in mAP and 1\% in Rank-1 accuracy, respectively. In the body mode, we compare with the strong method BoT~\cite{luo2019bag} and the SOTA method TransReid~\cite{He_2021_ICCV}. Table\,\ref{tab:repre-res-fbv} shows that URN is quite competitive compared with them. In the voice mode, compared with PSLA~\cite{gong2021psla}, Titanet~\cite{koluguri2021titanet}, and IDM~\cite{chung2020in}, URN has a significant advantages. It achieves satisfactory performance in three modalities, demonstrating the better feasibility of unified representation for multi-modal signals .

\textbf{Retrieval on Public Datasets}. 
Besides MTCG, we further validate URN on other public datasets, \emph{i.e.,} MegaFce~\cite{kemelmacher2016megaface}, Market-1501~\cite{zheng2015scalable}, and Voxceleb1~\cite{Nagrani17}. As shown in Table\,\ref{tab:repre-res-public}, the overall performance trend keeps consistent with the results on MTCG. For the body and voice modalities, our URN behaves better than other methods. For the face modality, our URN is still competitive. Note that MageFace is a large-scale face dataset, which contains 1M photos for distraction. Compared with Table\,\ref{tab:repre-res-fbv} and Table\,\ref{tab:repre-res-public}, our URN achieves satisfactory performances in large- and small-scale datasets, verifying its robust generalization.

\textbf{Intra-class Similarities and Inter-class Variances}. 
We further evaluate our unified representation in Fig.\,\ref{fig:similarity}. It visualizes the confusion matrix maps between features extracted from VGGFace2~\cite{cao2018vggface2} method and from our URN. For VGGFace2, many samples share similarities with the samples outside the red box (outside the same cluster). Instead, the sample pairs with high similarity in our map are much more compact inside the red box than outsides. This diagonal distribution proves that our features have higher intra-class similarities and inter-class variances.

%
%

\section{Conclusion}
This paper proposes a Unified and Dynamic Graph (UniDG) framework for temporal character grouping in long videos. Firstly, UniDG provides representation for different modalities by a unified representation network. Besides, it further proposes to leverage dynamic graphs for clustering. Next, it takes a progressive association method to fuse the multi-modal clustering results. 
Finally, to evaluate our proposed method, we collected by far the largest multi-modal dataset named MTCG. Extensive experiments on MTCG and public datasets demonstrate the effectiveness and generalization ability of our proposed method.





%
%

\ifCLASSOPTIONcaptionsoff
  \newpage
\fi





{
\bibliographystyle{IEEEtran}
\bibliography{egbib}
}

\end{document}